%% file: main.tex
\pdfoutput=1

\PassOptionsToPackage{dvipsnames}{xcolor}
\documentclass[11pt]{article}

\usepackage[final]{acl}

\usepackage{times}
\usepackage{latexsym}

\usepackage[T1]{fontenc}

\usepackage[utf8]{inputenc}

\usepackage{microtype}

\usepackage{inconsolata}

\usepackage{graphicx}

\usepackage{algorithm}
\usepackage{algorithmic}
\usepackage{color}
\usepackage{newfloat}
\usepackage{listings}
\usepackage{caption}
\usepackage{float} 
\usepackage{subfigure}
\usepackage{amsmath}
\usepackage{mathrsfs}
\usepackage{amsthm}
\usepackage{booktabs}
\usepackage{multirow}
\usepackage{amssymb}
\usepackage{bbding}
\usepackage[normalem]{ulem}
\usepackage{authblk}
\usepackage{fancyhdr}
\usepackage{url}

\title{MPL: Multiple Programming Languages with Large Language Models \\for Information Extraction}

\author{
Bo Li\textsuperscript{1},
Gexiang Fang\textsuperscript{2},
Wei Ye\textsuperscript{2}\textsuperscript{\textasteriskcentered},
Zhenghua Xu\textsuperscript{1}\thanks{Co-corresponding authors: \texttt{wye@pku.edu.cn}, \texttt{zhenghua.xu@hebut.edu.cn}},
Jinglei Zhang\textsuperscript{2},
Hao Cheng\textsuperscript{3},
Shikun Zhang\textsuperscript{2}
\\
\textsuperscript{1}State Key Laboratory of Intelligent Power Distribution Equipment and System, \\School of Health Sciences and Biomedical Engineering, Hebei University of Technology \\
\textsuperscript{2}National Engineering Research Center for Software Engineering, Peking University \\
\textsuperscript{3}School of Artificial Intelligence, Hebei University of Technology \\
\texttt{deepblue.lb@gmail.com, wye@pku.edu.cn, zhenghua.xu@hebut.edu.cn}
}

\begin{document}
\maketitle
\begin{abstract}
Recent research in information extraction (IE) focuses on utilizing code-style inputs to enhance structured output generation. The intuition behind this is that the programming languages (PLs) inherently exhibit greater structural organization than natural languages (NLs). This structural advantage makes PLs particularly suited for IE tasks. Nevertheless, existing research primarily focuses on Python for code-style simulation, overlooking the potential of other widely-used PLs (e.g., C++ and Java) during the supervised fine-tuning (SFT) phase. In this research, we propose \textbf{M}ultiple \textbf{P}rogramming \textbf{L}anguages with large language models for information extraction (abbreviated as \textbf{MPL}), a novel framework that explores the potential of incorporating different PLs in the SFT phase. Additionally, we introduce \texttt{function-prompt} with virtual running to simulate code-style inputs more effectively and efficiently. Experimental results on a wide range of datasets demonstrate the effectiveness of MPL. Furthermore, we conduct extensive experiments to provide a comprehensive analysis. We have released our code for future research~\footnote{https://github.com/PKU-Fgx/MPL}.
\end{abstract}

\input{1.intro}

\input{3.Method}

\input{4.experiments}

\input{5.results}

\input{2.related_work}

\section{Conclusion} 

In this paper, we propose MPL, a novel framework that leverages multiple programming languages with large language models for information extraction. We first propose \texttt{function-prompt} with virtual running component, a more lightweight and effective code-style simulate method compared to \texttt{class-prompt}. Then, MPL reformulates a given IE task and textual input into a code-style input with three programming languages, i.e., Python, C++ and Java to explore the potential of cooperation between them. Extensive experimental results on various IE datasets not only validate the effectiveness of MPL but also provide a thorough analysis to help readers better understand our method. To facilitate future research, we will release our code and necessary files later.

\section*{Limitation}
Although our research explores the potential of using multiple programming languages and achieves significantly better performance than previous works, training efficiency remains a notable drawback, leading to more than twice the training cost compared to the single-PL setting. Future work will focus on reducing training time under the MPL framework while maintaining performance.

\section*{Acknowledgements}
This work was supported by the National Natural Science Foundation of China (Grant No. 62276089), the Natural Science Foundation of Tianjin (Grant No. 24JCJQJC00200), the Natural Science Foundation of Tianjin (Grant No. 24JCQNJC01230), the Natural Science Foundation of Hebei Province (Grant No. F2024202064), the Science Research Project of Hebei Education Department (Grant No. BJ2025004), the Ministry of Human Resources and Social Security of China (Grant No. RSTH-2023-135-1), and the Science and Technology Program of Hebei Province (Grant No. 24464401D).

\bibliography{custom}
\clearpage
\input{appendix.tex}

\end{document}

%% file: 1.intro.tex
\section{Introduction}

Large language models (LLMs) \cite{DBLP:conf/nips/BrownMRSKDNSSAA20,hoffmann2022training,smith2022using,ouyang2022training,touvron2023llama} trained on massive natural language have demonstrated their superiority in a wide range of natural language processing tasks, such as machine translation \cite{li2023self,zhang2023prompting} and dialogue systems \cite{DBLP:conf/aaai/YangZZZXJZ24,nananukul2024if}. Nevertheless, researchers have found that LLMs do not achieve desirable performance on information extraction tasks, such as named entity recognition and relation extraction \cite{li2023evaluating,xu2023large,han2023information}. Since IE requires extracting structured knowledge from complex and variably expressed natural language, relying solely on natural language instructions, even with tailored prompts, may not be sufficient for generating structured outputs effectively \cite{li2023codeie,guo2023retrieval,li2024knowcoder}.

\begin{figure}[t]
	\centering
	\includegraphics[width=0.99\linewidth]{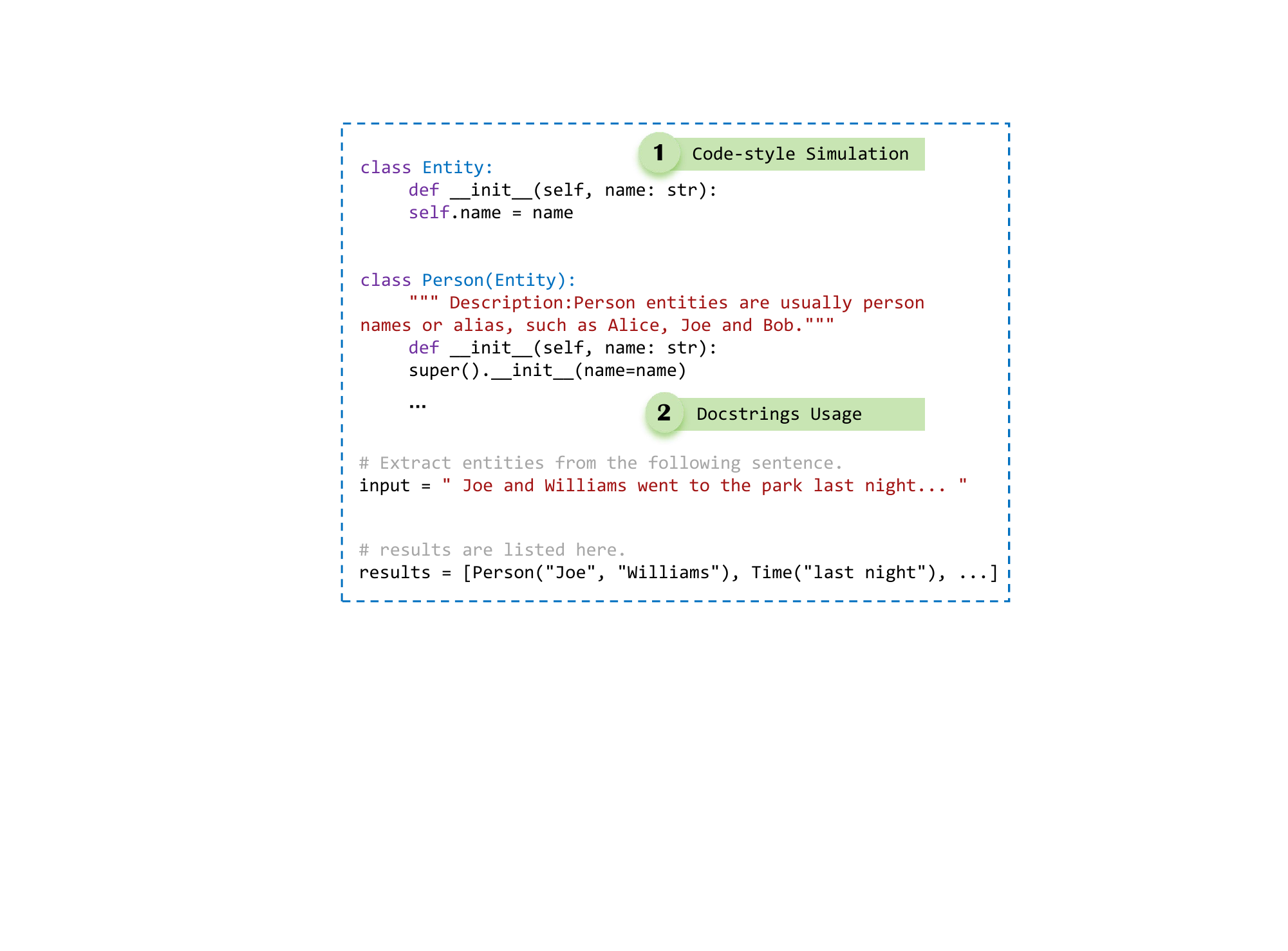}
	\caption{The typical procedure of code-style information extraction system, which mainly contains two components: the code-style simulation and the docstring usage.}
	\label{pic:intro}
\end{figure}

Recent studies have increasingly focused on exploring code-style input for IE tasks, as programming languages (PLs) are more formal and structured than natural languages (NLs) \cite{van2004concepts,dunn2019natural}, making them better suited for structured generation tasks. Broadly speaking, existing works \cite{wang2022code4struct,guo2023retrieval,li2023codeie,sainz2023gollie,bi2024codekgc,li2024knowcoder} mainly involve two components: \textbf{1) Code-style Simulation}—where the textual input is reformulated into a code-style representation using approach like \texttt{class-prompt}, which includes \texttt{base class} and \texttt{subclass} to represent the main task and fine-grained labels, along with additional elements to maintain standard code formatting. \textbf{2) Docstring Usage}—which involves using \texttt{docstrings} or \texttt{guidelines} for two main purposes: adding task and label descriptions within each class definition and providing explicit instructions for the task's inputs and outputs. The above procedures are briefly shown in Figure \ref{pic:intro}. Overall, existing works formulate IE tasks as code-style inputs, enabling LLMs to generate structured outputs more effectively and achieve strong performance in IE tasks.

Although \texttt{class-prompt} is commonly used to reformulate a given IE task into code-style input, it often requires defining numerous repetitive elements to maintain a standard format, such as the constructor \texttt{\_\_init\_\_} and the attribute assignment \texttt{self.name = name}. Besides, some label descriptions are repeated multiple times in \texttt{docstrings} due to the structural requirements. These redundancies not only result in verbose code but also significantly increase input length, negatively impacting both performance and training efficiency. To address these challenges, we propose \texttt{function-prompt} with a virtual running component, a more lightweight approach that wraps the task definitions, label descriptions and textual input into a single \texttt{function}, allowing LLMs generate the outputs by virtually executing the above \texttt{function}.

Additionally, previous work on code-style IE systems has primarily focused on Python, overlooking the multilingual nature of programming languages. Specifically, a wide range of commonly used PLs possess distinct inherent features, that could make them more suitable for specific IE tasks. For instance, Python's clean and readable syntax \cite{sarkar2016text,holst2021importance} facilitates handling complex nested structures and enhances clarity, making it a standard choice for code-style in IE systems. However, Python’s dynamic typing can lead to ambiguous types, which may introduce errors that are only detected during runtime. In contrast, C++ enforces strong typing, requiring explicit declarations of variable types \cite{stroustrup2013c++,stroustrup2014programming}, which helps prevent type-related errors early in development. Additionally, Java’s object-oriented nature \cite{blanchet1999escape,poo2008object} simplifies the representation of complex data structures, demonstrating the advantages of considering a variety of languages for diverse IE needs. In this paper, we propose \textbf{M}ultiple \textbf{P}rogramming \textbf{L}anguages with large language models for information extraction (shortened as \textbf{MPL}), a novel perspective that explores the integration of different PLs in the code-style IE system. To be specific, for a given textual input, we use three programming languages to simulate the coding process, i.e., Python, C++ and Java. These PLs are chosen for their common use and distinct characteristics, and the code-style inputs will be used in the supervised fine-tuning (SFT) process upon LLMs, aiming at exploring the cooperation between different PLs when dealing with IE tasks.

We conduct comprehensive experiments across a variety of IE tasks to evaluate the effectiveness of MPL. Specifically, we use widely used IE datasets for named entity recognition, relation extraction and event extraction to create datasets for SFT in the code-style format. We find that with the help of the simple and efficient \texttt{function-prompt}, MPL achieves significantly better performance compared to previous works that relied on larger model architectures or extensive external pre-training datasets. Additionally, our extensive experimental analysis of code-style simulations, LLM usage, and complementarity between different PLs provides valuable insights, serving as practical guidelines for future research. To sum up, our main contributions are as follows:

\begin{itemize}
    \item We are the first to utilize multiple programming languages with LLMs for IE tasks, aiming to explore the potential of combining different PLs. MPL's strategy involves generating various code-style inputs from a single textual input and performing SFT on LLMs. Additionally, we developed the \texttt{function-prompt} with virtual running for more efficient and effective code-style simulations.
    \item We carry out thorough experiments and analysis using diverse IE datasets, where MPL consistently achieves strong outcomes. Our experiments also extend to exploring different aspects of code-style simulation, LLM selection, and the interaction between various PLs, providing valuable insights for ongoing research.
\end{itemize}


%% file: 3.Method.tex
\begin{figure*}[t]
    \centering
    \includegraphics[width=0.99\linewidth]{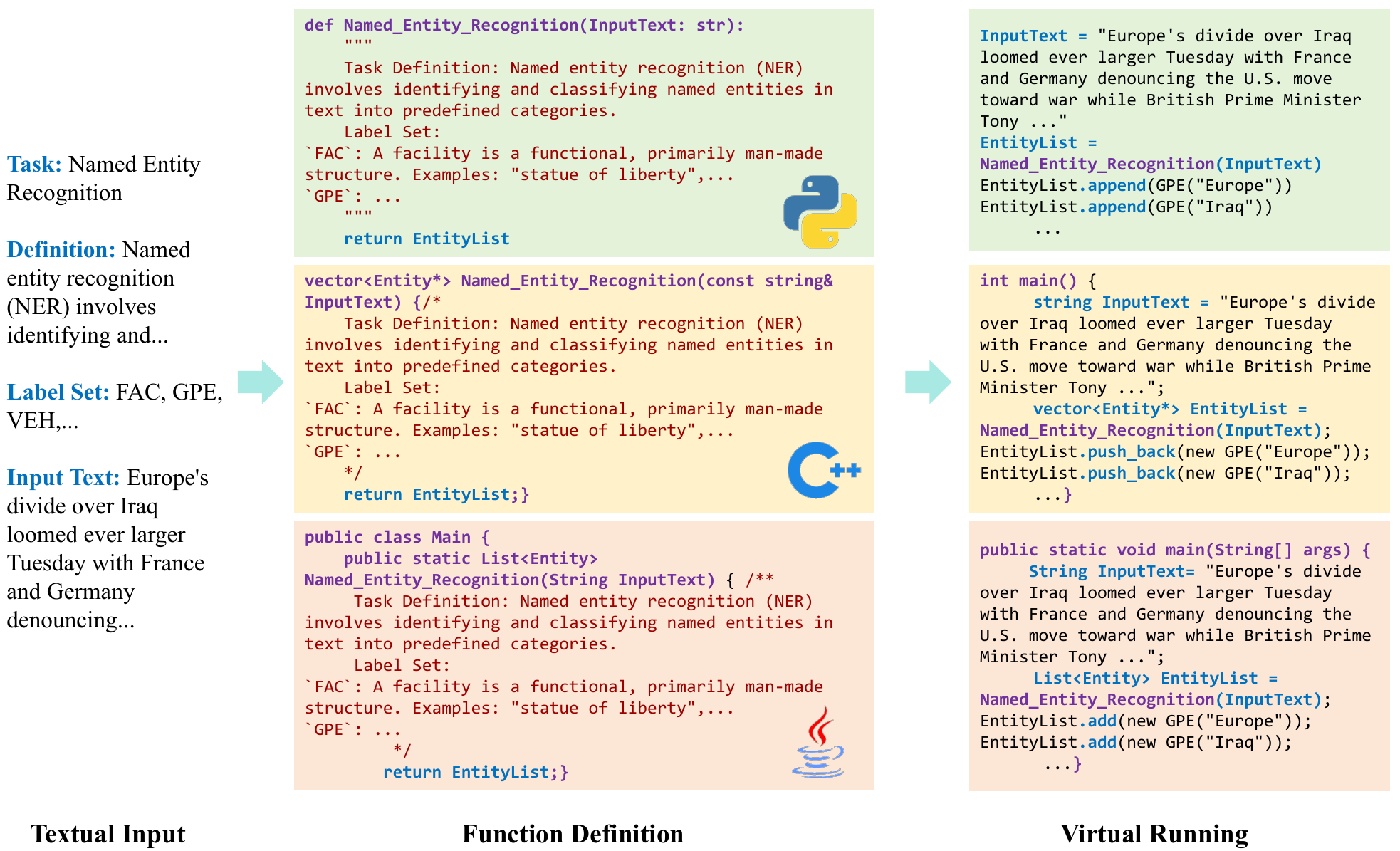}
    \caption{Our framework utilizes multiple programming languages, i.e, Python, C++, and Java, to convert elements from IE tasks and target textual inputs into code-style formats. To enhance the simulation process and help LLMs in processing textual inputs and generating outputs more naturally, we introduce the \texttt{function-prompt} with function definition and virtual running components. Better viewed in color.}
    \label{pic:model}
\end{figure*}

\section{Proposed Method}

In this paper, we propose \textbf{MPL}, a new framework that leverages multiple programming languages with large language models for information extraction. First, we briefly introduce the commonly used \texttt{class-prompt}, and then we propose a new design named \texttt{function-prompt}, which simulates the coding process in a more lightweight and effective way. Next, we introduce how to transform textual input into code-style representations across multiple PLs. Finally, we provide training details to clarify the implementation.

\subsection{\texttt{Class-prompt}}

The \texttt{Class-prompt} comprises two hierarchical levels: \texttt{base class} and \texttt{subclass}, representing the label definitions of a given IE task at different granularities. An overview is shown in Figure \ref{pic:intro}. The \texttt{Base class} serves as the parent class of all fine-grained entity, relation and event labels, denoted as \texttt{Entity}, \texttt{Relation} and \texttt{Event}. For instance, all entity labels are grouped under the \texttt{Entity} base class. Typically, the \texttt{base class} contains a constructor \texttt{\_\_init\_\_} and the attribute assignment \texttt{self.name = name}. The \texttt{subclass} utilizes \texttt{class} inheritance to represent more specific labels, such as a PERSON entity type being a \texttt{subclass} of \texttt{Entity}, defined as \texttt{class PERSON(Entity)}. Some approaches also incorporate event triggers within the corresponding \texttt{event subclass} when applicable. Besides, \texttt{docstrings} or \texttt{guidelines} are often used within \texttt{subclass} to wrap the task definitions and label descriptions, providing necessary information for understanding the given IE task and fine-grained labels easily and accurately.



\subsection{\texttt{Function-prompt}}

Although \texttt{class-prompt} effectively reformulates IE tasks into a standard code-style representations, it requires defining numerous repetitive elements to maintain the standard code format, and label descriptions often need to be repeated multiple times in \texttt{docstrings} or \texttt{guidelines} due to the nature of programming languages. These drawbacks significantly increase the length of the code-style inputs, negatively impacting both performance and training efficiency. To address these issues, this paper introduces \texttt{function-prompt}, designed to simplify the coding process while preserving the structural benefits of code-style input. The overall framework is shown in Figure \ref{pic:model}. Specifically, \texttt{function-prompt} mainly involves the following two parts:

\textbf{Function Definition.} Taking NER as an example, we first define a \texttt{function} that takes \texttt{InputText} as an input \texttt{variable}. Next, we add the task definition and label descriptions in the \texttt{docstring}, providing crucial information for understanding the given task and candidate labels. Specifically, we use ``Task Definition" and ``Label Set" to clearly highlight the target task and its corresponding candidate labels. At the end of the \texttt{function}, a \texttt{return} statement outputs the prediction, making our code-style input both complete and more reflective of actual coding practices. Besides, compared to the \texttt{class-prompt}, our \texttt{function-prompt} eliminates the need for repetitive additional elements and label descriptions, thereby providing a simpler and more lightweight method for simulating code-style inputs.


\textbf{Virtual Running.} Unlike previous approaches that solely rely on \texttt{docstrings} to guide models in processing inputs and generating outputs—a method that is somewhat unnatural for code-style inputs—our \texttt{function-prompt} incorporates a virtual running step. This step involves taking the textual input as input to the \texttt{function} and executing it virtually. Virtual running not only makes interaction with LLMs more intuitive under the code-style representation but also enhances their ability to generate structured outputs efficiently. During the generation phase, outputs are stored and displayed using a \texttt{list}. For instance, with the NER task, the \texttt{Named\_Entity\_Recognition} function runs virtually, and each identified entity is subsequently appended to an \texttt{EntityList}.


\subsection{Multiple Programming Languages for Information Extraction}

Instead of relying solely on Python, this study employs a broader range of programming languages, including Python, C++, and Java, to simulate the code-style inputs. MPL represents a given textual input from multiple programming languages perspectives and learns complementary information. For a given task and a textual input, we show different code-style inputs in Figure \ref{pic:model}, and the whole input for each programming language is provided in the Appendix C. 


To convert a given IE task and textual input into a code-style representation, we develop transforming functions for each programming language. These functions reformulate the textual input to align with the syntax and structure of the respective programming languages. In this process, we follow two key guidelines: 1) ensuring that all necessary information, such as the task name and definition, candidate labels and descriptions, along with the virtual running component, is wrapped within the chosen programming language; 2) maintaining consistency in the logic of the code design and the execution flow across Python, C++, and Java. Following these principles, we generate code-style inputs that convey similar semantic meanings while adopting diverse formats for IE tasks.


\subsection{Training and Testing}
During the training phase, these code-style inputs are used in the supervised fine-tuning (SFT) phase for IE tasks. We apply the standard Next Token Prediction loss, training our models solely on the actual output tokens while excluding the input sequences. For example, in the NER task shown in Figure \ref{pic:model}, we compute the loss only after \texttt{EntityList = }, and the output order follows the label definition order. All the code-style inputs with different programming languages are randomly shuffled before feeding into the model. During the testing phase, unless specified otherwise, inputs from three programming languages are used to derive predictions, which are then aggregated through a voting mechanism to determine the final results \footnote{We also evaluate the model using a single PL during testing and observed only a slight performance degradation, please refer Section \ref{sec:ablation} for more details.} (referred to as the default setting). 



%% file: 4.experiments.tex
\section{Experimental Setup}

\begin{table*}[]
\centering
\setlength{\tabcolsep}{0.7mm}
\begin{tabular}{c|cc|cc|cccc}
\toprule[1.5pt]
\textbf{}            & \multicolumn{2}{c|}{\textit{CodeLLaMA-7B}} & \multicolumn{2}{c|}{\textit{CodeLLaMA-13B}} &                   &                    &                     &                        \\ \hline
\textbf{}            & \textbf{GoLLIE}    & \textbf{MPL}          & \textbf{GoLLIE}        & \textbf{MPL}       & \textbf{SLM} & \textbf{KnowCoder} & \textbf{GoLLIE-34B} & \textbf{MPL-8B} \\ \midrule[1.5pt]
\textbf{ACE05-NER}   & 88.1               & 89.6                  & 89.4                   &   90.6           & 86.6              & 86.1               & 89.6                & 91.4                   \\ 
\textbf{BC5CDR}      & 87.5               & 87.9                  & 87.9                   &   87.8            & 91.9              & 89.3               & 88.4                & 89.6                   \\ 
\textbf{CoNLL03}     & 92.8               & 93.1                  & 93.0                   &   93.2             & 93.0              & 95.1               & 93.1                & 93.5                   \\ 
\textbf{DIANN}       & 79.4               & 84.8                  & 82.6                   &   85.7             & 74.8              & 94.7               & 84.1                & 85.4                   \\ 
\textbf{NCBID}        & 85.4               & 87.2                  & 86.5                   &  88.4              & 90.2              & 83.8               & 85.8                & 88.1                   \\ 
\textbf{OntoNotes5*} & 83.4               & 83.7                  & 84.0                   &  84.7             & 84.6              & -                  & 84.6                & 85.7                   \\ 
\textbf{WNUT2017}    & 52.0               & 53.1                  & 50.5                   &  52.6              & 60.2              & 66.4               & 54.3                & 52.6                   \\ 
\textbf{ACE05-RE}    & 63.6               & 65.8                  & 67.5                   &   66.0             & 66.1              & 64.5               & 70.1                & 70.8                   \\ 
\textbf{ACE05-EAE}    & 66.0               & 71.2                  & 67.8                   &   72.9             & 54.8              & 70.3               & 68.6                & 72.8                   \\ 
\textbf{RAMS}        & 48.7               & 48.9                  & 49.6                   &   49.4           & 48.6              & -                  & 51.2                & 50.9                   \\ 
\textbf{ACE05-EE}   & 72.2               & 70.7                  & 70.9                   &    70.1            & 73.4              & 74.2               & 71.9                & 72.7                   \\ \hline
\textbf{Avg.Score}        & 74.4               & \textbf{76.1(+1.7)}   & 75.4                   &    \textbf{76.5(+1.1)}            & 74.9              & -                  & 76.5                & \textbf{77.6(+1.1)}    \\ \bottomrule[1.5pt]
\end{tabular}
\caption{We conducted extensive experiments on CodeLLaMA-7B , CodeLLaMA-13B and LLaMA3-8B, where MPL-8B denotes the backbone is LLaMA3-8B. The values in brackets represent the performance differences between our MPL approach and the corresponding GoLLIE model. To minimize randomness, we tested each model five times and reported the average performance. Additionally, we conducted t-tests to compare our results with previous results, confirming that our results are statistically significant with a $p$-value of less than 0.05.}
\label{tab:main}
\end{table*}

\subsection{Dataset}\label{sec:data}

In this paper, we use the following widely used datasets across various IE tasks. These tasks include Named Entity Recognition (NER), Relation Extraction (RE), Event Argument Extraction (EAE) and Event Extraction (EE). Specifically, we use ACE05-NER\footnote{https://catalog.ldc.upenn.edu/LDC2006T06}, BC5CDR~\cite{wei2016assessing}, DIANN~\cite{fabregat2018overview}, NCBID~\cite{dogan2012improved}, WNUT2017~\cite{derczynski2017results}  CoNLL03 \cite{sang2003introduction},  and OntoNotes 5.0 \cite{pradhan2013towards} for NER, ACE05-RE for RE. As for event-related tasks, we use ACE05-EAE and RAMS \cite{ebner2019multi} for EAE, and ACE05-EE for EE. The above datasets cover a wide range of topics and are suitable for the SFT phase. Consistent with previous works \cite{sainz2023gollie,li2024knowcoder}, the Micro-F1 score is used to evaluate performances. The star (*) is used to indicate that the OntoNotes 5 dataset was aligned to the GOLLIE  \cite{sainz2023gollie} setup, which differs significantly from that used by KnowCoder in both training and testing. 



\subsection{Training Details}

We use the base version of StarCoder-v2 \cite{lozhkov2024starcoder}, LLaMa2-7B \cite{touvron2023llama}, CodeLLaMA-7B, CodeLLaMA-13B\cite{roziere2023code} and LLaMA3-8B \footnote{https://ai.meta.com/blog/meta-llama-3/} in our experiments~\footnote{We also use Qwen2.5-7B~\cite{yang2024qwen2} and QwenCoder2.5-7B~\cite{hui2024qwen2} to further test MPL's generalization ability with different LLM backbones, and the results are reported in Appendix E.}. We use QLoRA \cite{dettmers2024qlora} with lora\_rank = 96 and lora\_alpha = 192 for efficient training, the dropout ratio is set to 0.1. The models were trained with batch size of 8 and a learning rate of 1e-4 with a cosine scheduler and 0.05 warmup. The maximum length of input is 2048, the training epoch is 5. 


\subsection{Compared Models}

We mainly compare our method with GoLLIE \cite{sainz2023gollie} and KnowCoder \cite{li2024knowcoder}, both of which have recently achieved SoTA results and involve supervised fine-tuning phase on LLMs. GoLLIE uses specific data processing and guideline expansion operations, while KnowCoder uses 33 specific domain IE datasets in SFT stage. Besides, KnowCoder includes a schema understanding phase that contains millions of instances across various external knowledge to further pre-train LLMs. We also compare MPL with previous data-specific supervised fine-tuning, which uses small language models as backbone networks, such as BERT~\cite{DBLP:conf/naacl/DevlinCLT19} and RoBERTa~\cite{DBLP:journals/corr/abs-1907-11692}, denoted as SLM. We present these reference models in the Appendix B.

Furthermore, we have designed several model variants to conduct ablation studies. Specifically, to determine whether simply increasing data size can yield significant improvements, we trained a model with a \textit{$3\times$} larger dataset based on python-style inputs, denoted as Python$_{3\times}$. We trained a model named as MPL$_{sampled}$, which randomly selects one PL instance per input, resulting in a dataset one-third the size of the full multiple PL dataset. MPL$_{sampled}$ aims to investigate if employing the same data size in multiple PL settings can achieve satisfactory performance. Additionally, we report the performance of models trained with a single programming language during the SFT phase for a detailed analysis.

%% file: 5.results.tex
\section{Results and Analysis}
We conducted extensive experiments to verify the effectiveness of MPL, below are our results and findings.



\begin{table*}[]
\centering
\setlength{\tabcolsep}{2mm}
\begin{tabular}{@{}c|cc|cc@{}}
\toprule[1.5pt]
                      & \multicolumn{2}{c|}{\textit{\textbf{Average Score}}} & \multicolumn{2}{c}{\textit{\textbf{Language Capability}}}                \\ \cmidrule(l){2-5} 
\textbf{}             & \textit{\textbf{Python}}   & \textit{\textbf{MPL}}   & \textit{\textbf{MMLU (NL)}} & \textit{\textbf{HumanEval Pass@1 (PL)}} \\ \midrule[1.5pt]
\textbf{StarCoder-v2} & 72.7                       & 73.8                    & 38.3                        & 35.4                                    \\
\textbf{LLaMa2-7B} & 75.9                       & 76.6                    & 45.3                        & 12.2                                    \\
\textbf{LLaMA3-8B}    & 76.4                       & 77.6                    & 66.7                        & 37.2                                    \\ \bottomrule[1.5pt]
\end{tabular}
\caption{The results of various backbone LLMs trained with Python and MPL. We also report the the \textit{natural language ability} (shortened as \textit{NL}) and \textit{programming language ability} (shortened as \textit{PL}) of each backbone LLMs. These results show that the natural language ability is more important than programming language ability in the case of information extraction tasks.}
\label{tab:backbone}
\end{table*}
\subsection{Main Results}\label{sec:main}
The main results are shown in Table \ref{tab:main}, from which we can see that: 1) MPL achieves the best overall performance compared to both SLM-based and LLM-based SoTA models. Notably, MPL-8B outperforms the largest GoLLIE model (34B parameters) by 1.1\% and surpasses the previous SLM-based SoTA by 2.7\%. Besides, MPL demonstrates competitive performance against KnowCoder, despite KnowCoder incorporates a schema understanding phase with millions of instances and uses 33 domain-specific IE datasets in the SFT stage. 2) For the same backbone, MPL consistently outperforms GoLLIE, achieving a 1.7\% improvement for CodeLLaMA-7B and 1.1\% for CodeLLaMA-13B. Moreover, MPL trained with CodeLLaMA-13B even achieves performance comparable to GoLLIE trained with the CodeLLaMA-34B model. These findings verify strength of MPL across various IE tasks, supporting both the motivation behind our research and the effectiveness of our model. 3) MPL consistently improves across different model configurations, including CodeLLaMA-7B, CodeLLaMA-13B and LLaMA3-8B. This suggests that MPL generalizes well across various model sizes. Its ability to deliver strong results with different backbones highlights its robustness and versatility for information extraction tasks.

\begin{table}[ht]
\centering
\begin{tabular}{lcc}
\toprule[1.5pt]
\textbf{Dataset} & \textbf{GoLLIE-34B} & \textbf{MPL} \\
\toprule[1.5pt]
WikiEvents-NER & 81.3 & \textbf{82.1} \\
Broad\_Twitter & 50.3 & \textbf{55.5} \\
MIT\_Movie & \textbf{62.4} & 62.2 \\
WikiEvents-EAE & \textbf{50.7} & 47.7 \\
\toprule[1.5pt]
\end{tabular}
\caption{Comparison of GoLLIE-34B and MPL across various zero-shot IE datasets.}
\label{tab:zero}
\end{table}

\subsection{Zero-shot Performance}

To assess the generalization capability of MPL beyond its training distribution, we evaluate it on four information extraction (IE) benchmarks that are excluded from the training set: WikiEvents-NER~\cite{li2021document}, Broad\_Twitter~\cite{derczynski2016broad}, MIT\_Movie~\cite{liu2013asgard} and WikiEvents-EAE~\cite{li2021document}. The results of MPL-8B and GoLLIE-34B are presented in Table~\ref{tab:zero}. On WikiEvents-NER, MPL achieves the highest score, despite having fewer parameters and undergoing lighter training. Even on datasets where GoLLIE-34B performs slightly better, the performance gap remains minimal. These findings indicate that MPL exhibits strong generalization across diverse datasets and task structures.

\subsection{Ablation Study}\label{sec:ablation}

To evaluate the effectiveness of proposed settings in MPL, we conducted a thorough ablation study, with results shown in Table~\ref{tab:ablation}. \textbf{1) Single vs. Multiple Programming Languages:} In the first block, we trained LLaMA3-8B model using a single PL with the \texttt{function-prompt}. The results reveal that performance with a single PL significantly lags behind MPL, highlighting the advantage of using multiple PLs for IE tasks. \textbf{2) Data Construction Methods:} In the second block, we explored different data construction methods. When we increased the Python-style dataset size threefold (Python$_{3\times}$), we observed severe overfitting and a notable drop in performance compared to MPL (-1.5\% on the average score). In contrast, randomly sampling one-third of the MPL training data resulted in a surprisingly competitive performance (-0.5\% decrement). These findings suggest that MPL's improvements stem from using multiple PLs with diverse representations, rather than merely increasing the dataset size. \textbf{3) Virtual Running Component:} In the final block, we evaluated the virtual running component, which simulates the coding process more naturally, and we found that it provided a 0.4\% improvement in performance. Detailed performance metrics are provided in Appendix A.

\begin{table}[H]
\centering
\setlength{\tabcolsep}{4mm}
\begin{tabular}{@{}c|c@{}}
\toprule[1.5pt]
\textbf{Training Setting} & \textbf{Average Score} \\ \midrule[1.5pt]
Python                    & 76.4(-1.2)               \\
C++                       & 76.6(-1.0)               \\
Java                      & 76.4(-1.2)               \\\hline
Python$_{3\times}$        & 76.1(-1.5)         \\
MPL$_{sampled}$           & 77.1(-0.5)         \\\hline
MPL                       & \textbf{77.6}      \\
w/o. Virtual Running      & 77.2(-0.4)         \\ \bottomrule[1.5pt]
\end{tabular}
\caption{Ablation studies on various components in MPL, we use LLaMA3-8B and \texttt{function-prompt} as default setting here.}
\label{tab:ablation}
\end{table}

\section{Analysis}

\subsection{Which Backbone Is More Suitable For Code-style Input?}\label{sec:backbone}

Intuitively, IE tasks are knowledge-intensive and require not only strong natural language (NL) understanding to capture semantic meaning but also robust programming language (PL) capabilities to generate structured knowledge. In this subsection, we evaluate which LLM backbone is most suited for code-style inputs in the context of IE tasks. The tested LLMs include StarCoder-v2, LLaMa2-7B, and LLaMA3-8B. For a deeper analysis, we roughly measure the language capabilities based on the performance in the MMLU~\cite{DBLP:conf/iclr/HendrycksBBZMSS21} (measuring NL ability) and HumanEval~\cite{DBLP:journals/corr/abs-2107-03374}  (measuring PL ability)~\footnote{The results of MMLU and HumanEval are cited from StarCoder-V2 Tech Report, LLaAa2 Tech Report and LLaAa3 Tech Report.}.

From Table \ref{tab:backbone} we can see that: 1) StarCoder-v2 performs the worst in IE tasks due to its limited natural language understanding capabilities, despite excelling in PL tasks. On the other hand, LLaMa2-7B achieves significantly better results, even with weaker PL capabilities. 2) LLaMA3-8B achieves the best result, owing to its superior NL and PL understanding, making it the ideal backbone for code-style inputs. In conclusion, the NL ability is more important than PL ability in the case of IE with code-style inputs. However, proper PL understanding ability could further enhance the overall performance.

\subsection{Disentangling Programming Language Diversity from Ensemble Effects}

To rigorously assess whether the performance gains of our MPL framework stem from true programming language diversity or from general ensemble effects, we designed a series of controlled baseline variants. These include: (1) training with a single programming language (Python-only); (2) ensembling multiple models trained with different random seeds using the same language; (3) prompt augmentation by varying label order or format while tripling data volume; and (4) ensembling independently trained models using different programming languages (Python, C++, Java). These variants isolate different sources of potential performance improvements, including random initialization effects, superficial prompt perturbations, and structural diversity introduced through programming language variation. Our unified MPL design incorporates multi-language supervision during training, aiming to embed more diverse code-level semantics into the model’s representation.

\begin{table}[H]
\centering
\begin{tabular}{l c}
\toprule[1.5pt]
\textbf{Model Variant} & \textbf{Avg. F1} \\
\toprule[1.5pt]
Single PL (Python only) & 76.40 \\
Seed Ensemble (Python × 3 seeds) & 76.60 \\
Label Reordering + 3× data & 76.22 \\
Prompt Format Variation + 3× data & 76.63 \\
PL Ensemble (Python, C++, Java) & 76.73 \\
MPL (ours) & \textbf{77.60} \\
\toprule[1.5pt]
\end{tabular}
\caption{Performance comparison of MPL with various ablation baselines.}
\label{tab:mpl_variants}
\end{table}

As summarized in Table~\ref{tab:mpl_variants}, ensembling models trained with different random seeds (76.60 F1) or applying prompt variations with expanded data (76.22–76.63 F1) yields only marginal gains over the single-language baseline (76.40 F1). In contrast, ensembling across different programming languages improves performance more noticeably (76.73 F1), suggesting that language diversity contributes beyond initialization noise or prompt form. Notably, our unified MPL training outperforms all variants (77.60 F1), achieving a +1.2 improvement over the Python-only baseline. These results confirm that programming language diversity introduces deeper structural and syntactic variability, fostering more robust task understanding and generalization. The observed advantage cannot be fully explained by general ensembling or superficial variations, validating the core design principle of MPL.

\begin{figure}[H]
    \centering
    \includegraphics[width=1.0\linewidth]{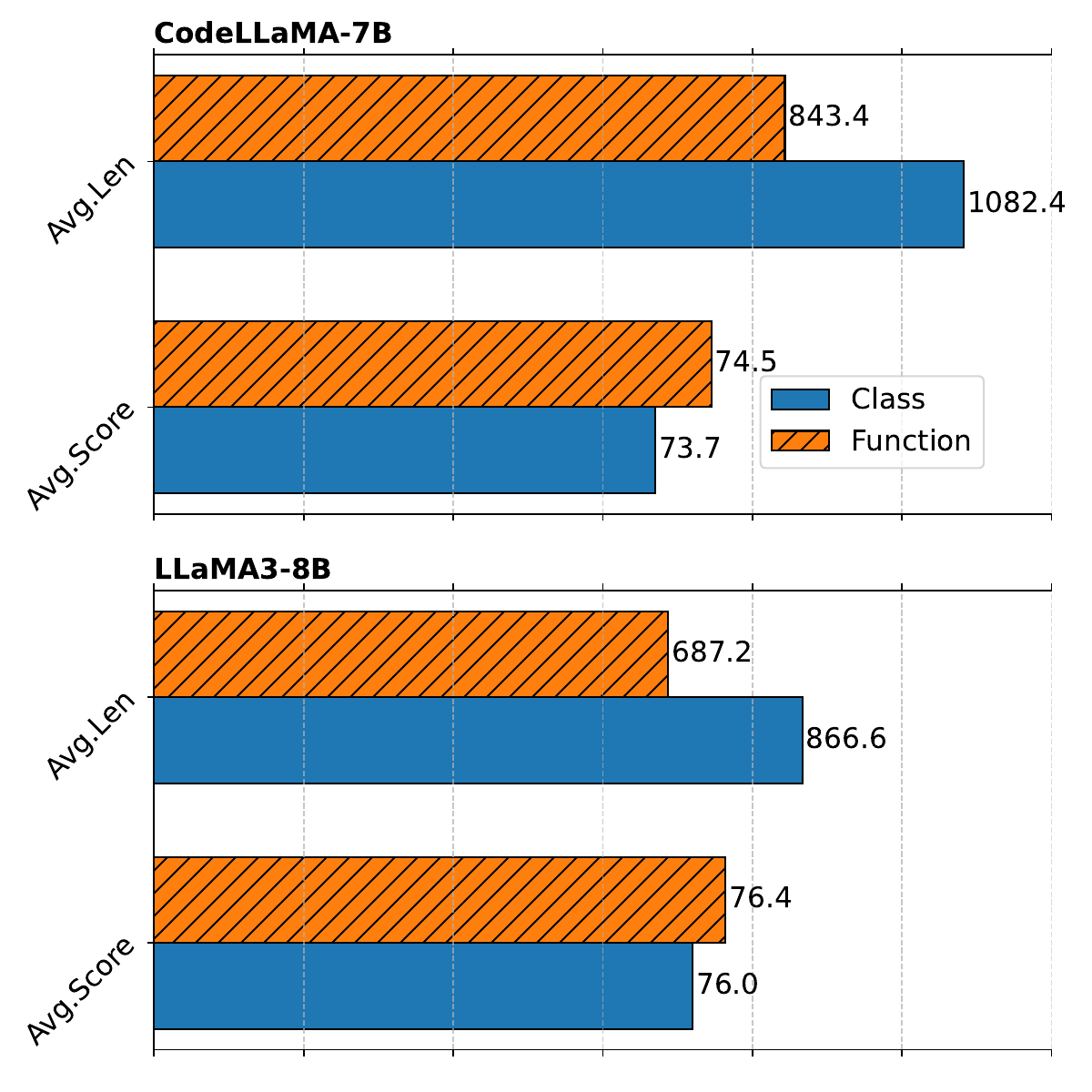}
    \caption{Performance and training statistics for different LLMs with various input formats. \textbf{Avg. Score} represents the average score across all datasets, while \textbf{Avg. Len} denotes the average input length after tokenization with the corresponding LLM's tokenizer. Notably, the same prompt yields different \textbf{Avg. Len} values across models due to variations in their tokenizer configurations. Detailed counts are provided in Appendix D}
    \label{pic:length}
\end{figure}

\subsection{\texttt{Function-prompt} vs. \texttt{Class-prompt}}\label{sec:prompt}

Another contribution of this research is the introduction of \texttt{function-prompt} as the alternative to \texttt{class-prompt}. In this subsection, we provided an in-depth analysis comparing the performance and training efficiency of these two prompts. Specifically, we report the average score and the average input length among all training datasets. Figure \ref{pic:length} presents the experimental results using different input formats. In terms of performance, our \texttt{function-prompt} consistently outperforms the \texttt{class-prompt} across various LLMs. Regarding training efficiency, the average input length of \texttt{function-prompt} is reduced around 20\% compared to \texttt{class-prompt}, highlighting its efficiency in processing inputs.


\subsection{Complementarity Between Different Programming Languages}\label{sec:diff_pl}

This subsection presents detailed performance analysis of different PLs and aggregation methods within our MPL framework. We investigate single PL predictions and two aggregation methods: \textbf{Voting} (Default), which  combines predictions from all three PLs to determine the final output, and \textbf{Union}, where a prediction is considered correct if any of the PLs produce the correct result, representing the union of all predictions. The results are shown in Table \ref{tab:union} and we can see that: 1) The performance benefit of \textbf{Voting} over individual PL predictions is minimal. This suggests that MPL effectively captures and integrates the diverse semantic meanings and formats inherent in different PLs, leading to robust performance. Consequently, selecting a single PL during testing can yield desirable results with reduced inference costs. 2) Aggregating results via \textbf{Union} significantly improves average scores, confirming our hypothesis that different PLs provide complementary strengths for varied IE tasks and datasets. Although \textbf{Voting} has been the default, these findings suggest exploring more effective aggregation methods could be beneficial. 

\begin{table}[H]
\centering
\setlength{\tabcolsep}{1.3mm}
\begin{tabular}{c|c|c}
\toprule[1.5pt]
\textbf{Model}                                                                         & \textbf{Prediction Setting} & \textbf{Avg.Socre} \\ \midrule[1.5pt]
\multirow{5}{*}{\textbf{\begin{tabular}[c]{@{}c@{}}MPL\\ (CodeLLaMA)\end{tabular}}} & Python                 & 75.6                  \\
                                                                                       & C++                    & 75.8                  \\
                                                                                       & Java                   & 75.4                  \\ \cline{2-3} 
                                                                                       & \textbf{Voting}       & 76.1                  \\
                                                                                       & \textbf{Union}         & \textbf{78.4(+2.3)}  \\ \hline
\multirow{5}{*}{\textbf{\begin{tabular}[c]{@{}c@{}}MPL\\ (LLaMa3)\end{tabular}}}    & Python                 & 77.2                  \\
                                                                                       & C++                    & 77.3                  \\
                                                                                       & Java                   & 77.3                  \\ \cline{2-3} 
                                                                                       & \textbf{Voting}        & 77.6                  \\
                                                                                       & \textbf{Union}         & \textbf{79.5(+1.9)}  \\ \bottomrule[1.5pt]
\end{tabular}
\caption{The average score with different prediction settings. The results are all obtained from well-trained MPL-8B models.}
\label{tab:union}
\end{table}

Besides, we compute the jaccard similarity coefficient \cite{jaccard1912distribution} to evaluate the similarities between different PLs' predictions. First, we compute the Jaccard similarity between each pair of PLs, and then we show the average coefficients in Table \ref{tab:jaccard}. We also analysis the performance gap between \textbf{Voting} and \textbf{Union} for a more comprehensive evaluation. From the results we can observe that: 1) For simple tasks like NER, the models perform well and the predictions across PLs are quite similar, showing minimal differences between ensemble methods. 2) For more complex tasks such as RE and EE, we not only observe a significant drop in overall model performance but also a substantial discrepancy in prediction similarity. This indicates that different languages excel at handling different tasks and datasets, leading to lower prediction similarity and more substantial performance differences between different ensemble methods. This finding strongly supports our motivation that different PLs have distinct characteristics and excel in different tasks. Through the design of MPL, we effectively harness the potential for cross-linguistic enhancement, thereby improving overall performance.

\begin{table}[h]
\centering
\setlength{\tabcolsep}{1mm}
\begin{tabular}{c|c|c|c}
\toprule[1.5pt]
\textbf{}                   & \textbf{\begin{tabular}[c]{@{}c@{}} Jaccard\\ Similarity\end{tabular}} & \textbf{MPL} & \textbf{Union-Voting} \\ \midrule[1.5pt]
\textbf{ACE05-NER}          & 0.98                        & 91.4         & 2.25                        \\ 
\textbf{BC5CDR}            & 0.99                        & 89.6       & 1.37                        \\ 
\textbf{CONLL03}      & 0.99                        & 93.5         & 1.45                        \\ 
\textbf{DIANN}      & 0.98                        & 85.4         & 0.08                        \\\hline
\textbf{ACE05-RE}           & 0.93                        & 70.8 & 16.25               \\ 
\textbf{ACE05-EAE}          & 0.94                        & 72.8 & 2.93               \\ 
\textbf{RAMS}               & 0.88                        & 50.9 & 3.55              \\ 
\textbf{ACE05-EE}           & 0.97                        & 72.7 & 3.67              \\ \bottomrule[1.5pt]
\end{tabular}
\caption{We calculate the jaccard similarity between each pair of PLs and report the average coefficients, the higher, the similar. We also show the performance gap between voting and union.}
\label{tab:jaccard}
\end{table}

%% file: 2.related_work.tex
\section{Related Work}
Information Extraction involves entity recognition, relation extraction, event extraction and other related tasks~\cite{li2024labels,vajjala2022we,li2023reviewing,li2023sequence,gao2024promptre,lu2021text2event,wang2022query}. There are plenty works focusing on information extraction in the era of large language model. Earlier research mainly evaluated the performances on various large language models under few-shot and zero-shot settings. For example, \citet{li2023evaluating,han2023information,li2024empirical} evaluated LLMs' performance and other dimensions to provide systemically analysis, while \citet{wadhwa2023revisiting,xu2023large,zhu2023llms,jiang2024genres} discussed the opportunities and challenges for IE field research in the era of large language model. 

Recent studies have explored the use of code-style inputs~\cite{chen2021evaluating,roziere2023code,li2023starcoder,zheng2023codegeex} for IE tasks, aiming to structure these tasks more formally and facilitate LLMs in generating structured outputs more accurately. CodeIE\cite{li2023codeie,wang2022code4struct} introduced code-style representations for various IE tasks under few-shot settings. Code4UIE\cite{guo2023retrieval} proposed a retrieval-augmented code generation framework, integrating class-prompting and multiple example retrieval strategies to enhance few-shot performance. CodeKCG~\cite{bi2024codekgc} reformulated natural language into code-style formats for generative knowledge graph construction, incorporating rationale-enhanced generation as an intermediate step to provide additional contextual information for unseen examples. Additionally, some studies have explored fine-tuning LLMs with code-style inputs. GoLLIE\cite{sainz2023gollie} introduced fine-grained label descriptions and candidate selection to improve supervised and zero-shot performance. KnownCoder\cite{li2024knowcoder} first designed a schema understanding phase with millions of pre-training instances to enhance LLM comprehension of code-style inputs, followed by a supervised fine-tuning phase on 33 domain-specific datasets, referred to as the schema-following phase.

%% file: appendix.tex
\section*{Appendix}
\subsection*{Appendix A: Detailed Results of Ablation Study}\label{app:ablation}
In Table~\ref{tab_app:ablation}, we provide the detailed results of each model variant in the ablation study.

\subsection*{Appendix B: The Reference Paper of Previous SoTA Method}\label{app:slm}

We compare our method with previous best SLM-based methods, these methods mainly build upon BERT and RoBERTa. We show the reference papers and the results in Table~\ref{table:slm}, and the results are the same as we reported in Table~\ref{tab:main} in the main paper.

\begin{table}[h]
\centering
\setlength{\tabcolsep}{1.5mm}
\begin{tabular}{@{}c|c|c@{}}
\toprule[1.5pt]
\textbf{DataSet} & \textbf{Result} & \textbf{Reference Paper}                     \\ \midrule[1.5pt]
ACE05-NER        & 86.6            & \citet{wang2023instructuie} \\
BC5CDR           & 91.9            & \citet{zhang2022optimizing} \\
CoNLL03          & 93.0            & \citet{lu2022unified}       \\
DIANN            & 74.8            & \citet{zavala2018hybrid}    \\
NCBID            & 90.2            & \citet{wang2023instructuie} \\
OntoNotes5       & 84.6            & \citet{sainz2023gollie}     \\
WNUT2017         & 60.2            & \citet{wang2021improving}   \\
ACE05-RE         & 66.1            & \citet{lu2022unified}       \\
ACE05-EAE        & 54.8            & \citet{lu2022unified}       \\
RAMS             & 48.6            & \citet{li2021document}      \\
ACE05-EE         & 73.4            & \citet{lu2022unified}       \\ 
\bottomrule[1.5pt]
\end{tabular}
\caption{The results of previous best SLM-based performances and their corresponding papers.}
\label{table:slm}
\end{table}

\subsection*{Appendix C: Detailed Inputs with Each Programming Language}\label{app:detailed_input}

\begin{figure*}[t]
    \centering
    \includegraphics[width=0.9\linewidth]{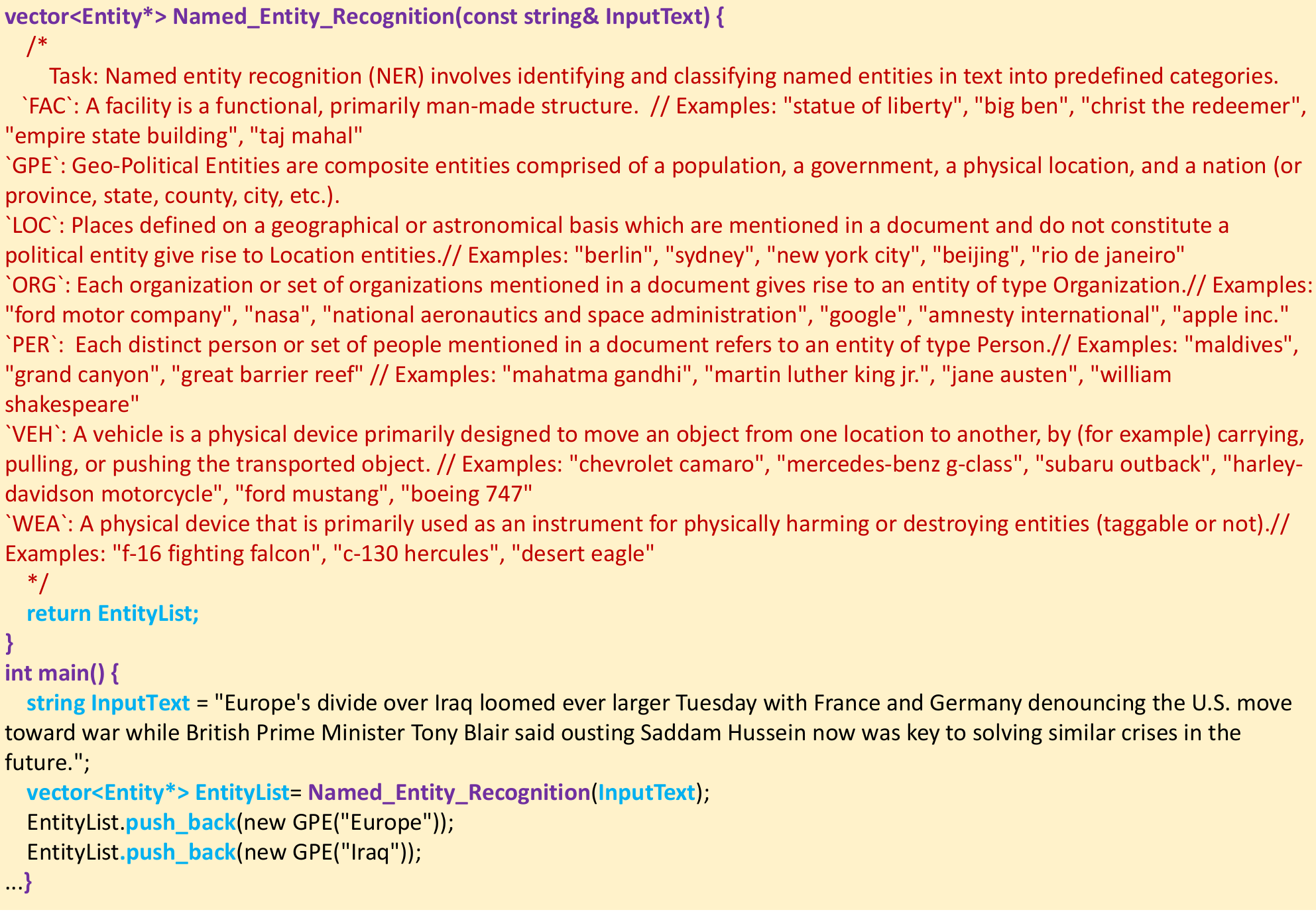}
    \caption{The detailed input using Python and \texttt{function-prompt} on ACE05-NER dataset.}
    \label{pic:app_python}
\end{figure*}

\begin{figure*}[t]
    \centering
    \includegraphics[width=0.9\linewidth]{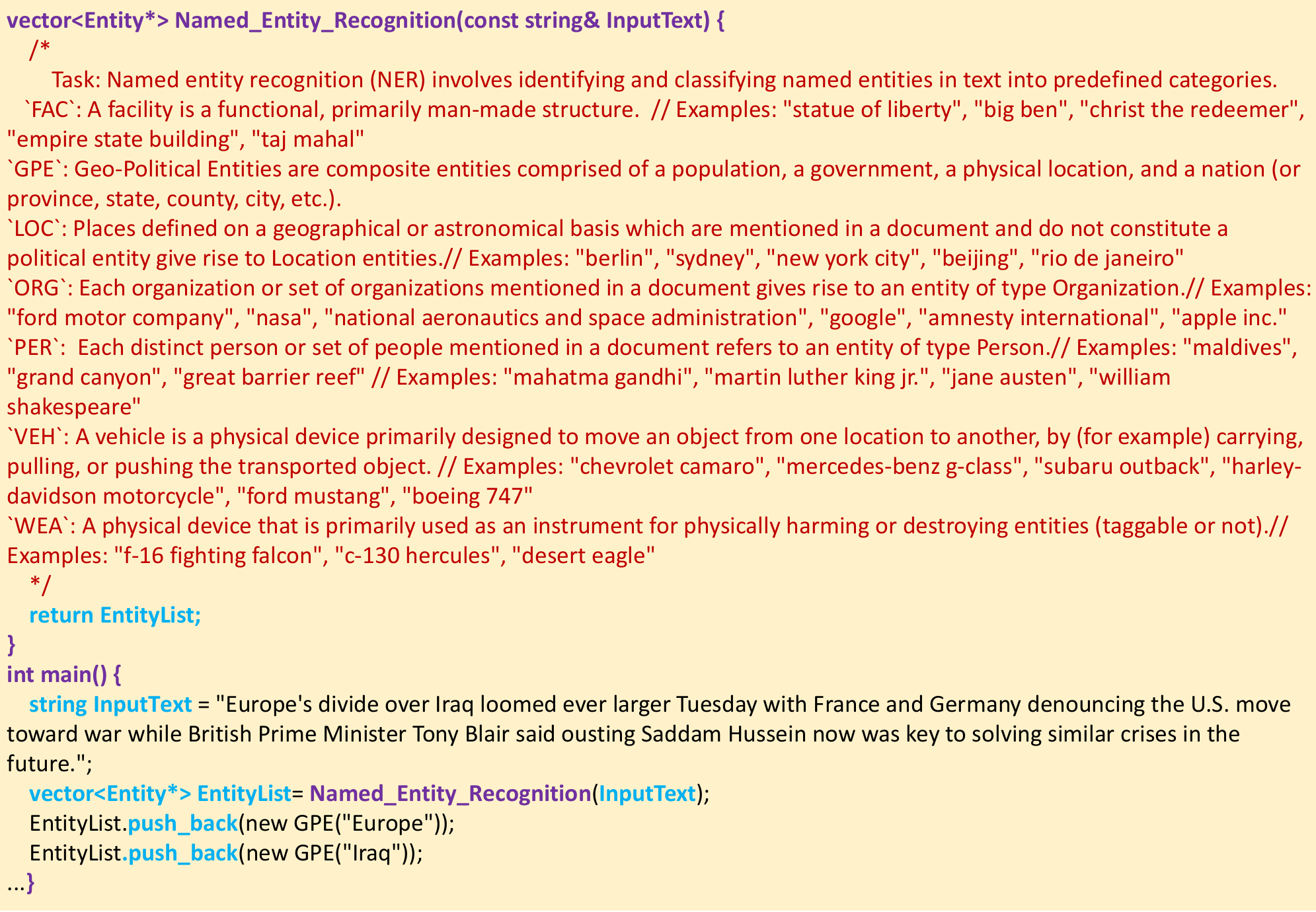}
    \caption{The detailed input using C++ and \texttt{function-prompt} on ACE05-NER dataset.}
    \label{pic:app_c}
\end{figure*}

\begin{figure*}[t]
    \centering
    \includegraphics[width=0.9\linewidth]{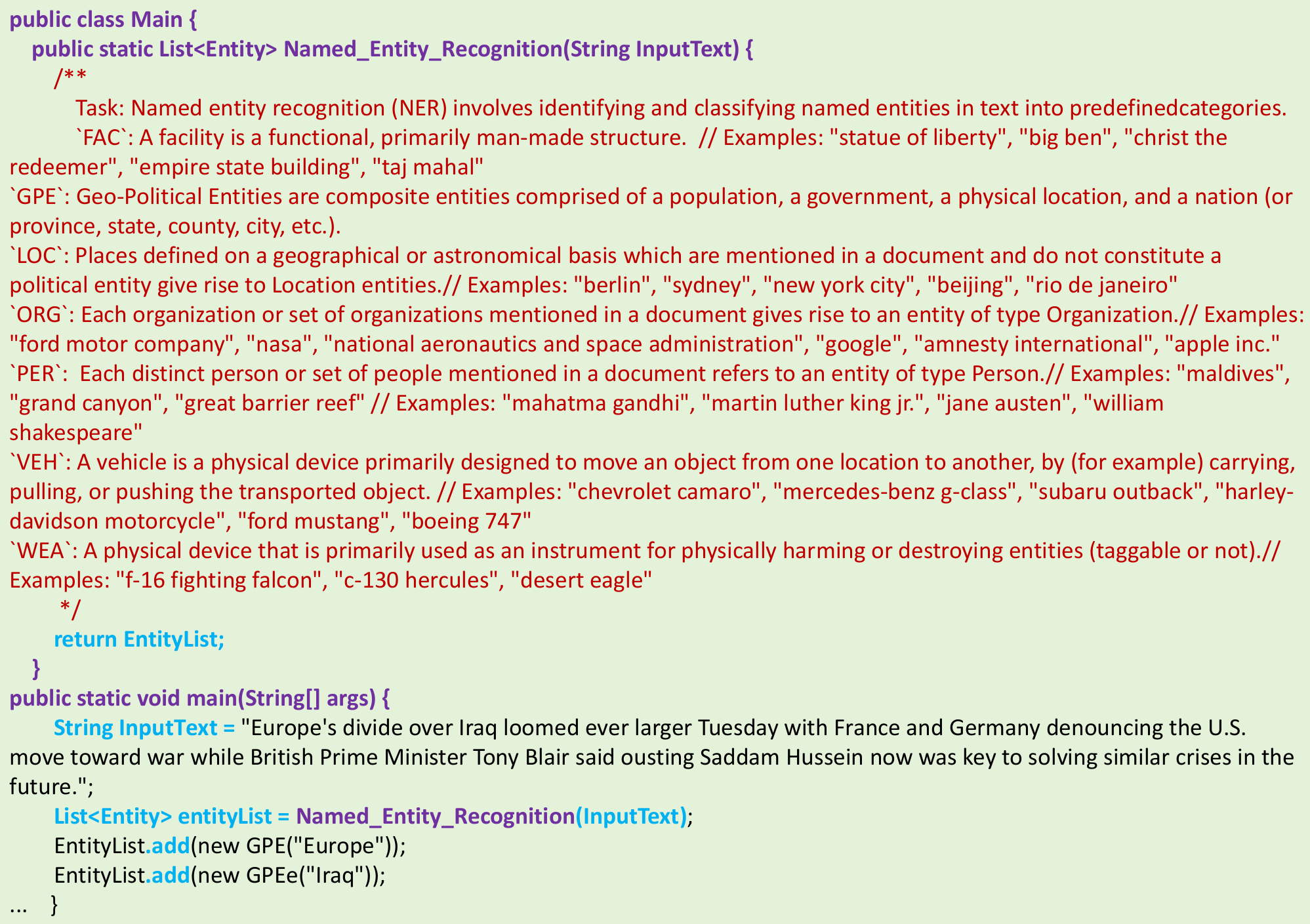}
    \caption{The detailed input using Java and \texttt{function-prompt} on ACE05-NER dataset.}
    \label{pic:app_java}
\end{figure*}

To help readers understand our method clearly and easily, we show the detailed input example for NER task in the ACE05 dataset using Python (Figure \ref{pic:app_python}), C++ (Figure \ref{pic:app_c}) and Java (Figure \ref{pic:app_java}). Please refer our submitted code file for inputs of whole dataset.

\subsection*{Appendix D: Sample Counts by Length Interval with Different Prompts}\label{app:length}

We list the detailed sample counts in different length intervals, this is the supplementary information for Figure~\ref{pic:length}.

Table \ref{app:table.length} shows that the \texttt{function-prompt} has a significant peak in the 300-600 length interval for around 49\% samples, indicating that a large majority of the function prompts are clustered within this length range. The 400-800 length interval is the most significant for \texttt{class-prompt}, accounting for approximately 52\% of all samples. This shows that \texttt{function-prompts} are heavily skewed towards shorter lengths, while \texttt{class-prompts} are generally longer.

\begin{table}[]
\centering
\setlength{\tabcolsep}{0.5mm}
\begin{tabular}{@{}c|c|c@{}}
\toprule[1.5pt]
\textbf{Length Interval} & \texttt{Class-prompt} & \texttt{Function-prompt} \\ \midrule[1.5pt]
0-99            & 0.00                  & 0.00                     \\
100-199         & 0.00                  & 0.02                     \\
200-299         & 4.74                  & 7.87                     \\
300-399         & 6.47                  & 19.11                    \\
400-499         & 22.43                 & 12.92                    \\
500-599         & 8.92                  & 17.24                    \\
600-699         & 16.45                 & 3.84                     \\
700-799         & 4.80                  & 0.70                     \\
800-899         & 1.31                  & 0.88                     \\
900-999         & 0.68                  & 2.37                     \\
1000-1099       & 0.10                  & 13.05                    \\
1100-1199       & 2.29                  & 15.19                    \\
1200-1299       & 1.31                  & 5.48                     \\
1300-1399       & 14.41                 & 1.28                     \\
1400-1499       & 7.01                  & 0.03                     \\
1500+       & 9.08                  & 0.00                     \\
\bottomrule[1.5pt]
\end{tabular}
\caption{Proportion of sample counts by length interval for \texttt{function-prompt} and \texttt{class-prompt}. We use LLaMa3-8B's tokenizer here.}
    \label{app:table.length}
\end{table}

\subsection*{Appendix E: Performances with Qwen2.5 Series.}\label{app:length}

To evaluate MPL's generalization ability across different LLM backbones, we conducted extensive experiments using MPL and its variants on Qwen2.5-7B\cite{yang2024qwen2} and QwenCoder2.5-7B\cite{hui2024qwen2}. The results, presented in Table~\ref{app:table.qwen}, show that models using a single programming language (e.g., Python) perform significantly worse than MPL, while MPL$_{smapled}$ shows slightly worse average score. These findings not only confirm the generalization ability of MPL, but also reinforce the key observations discussed in the main paper.

\begin{table*}[]
\centering
\setlength{\tabcolsep}{2.0mm}
\begin{tabular}{c|ccc|ccc}
\toprule[1.5pt]
\textbf{}            & \multicolumn{3}{c|}{\textit{Qwen2.5-7B}} & \multicolumn{3}{c}{\textit{Qwen Coder2.5-7B}}           \\ \hline
\textbf{}            & \textbf{Single PL}    & \textbf{MPL$_{smapled}$}          & \textbf{MPL}        & \textbf{Single PL}       & \textbf{MPL$_{smapled}$} & \textbf{MPL} \\ \midrule[1.5pt]
\textbf{ACE05-NER}   & 90.4               & 91.0                  & 91.2                   &   90.7           & 91.2              & 91.5  \\ 
\textbf{BC5CDR}      & 90.1               & 90.1                  & 90.2                   &   89.8            & 90.3              & 90.4   \\ 
\textbf{CoNLL03}     & 93.9               & 93.9                  & 93.4                   &   93.7             & 94.1              & 93.6  \\ 
\textbf{DIANN}       & 85.2               & 85.0                  & 84.1                   &   83.6             & 85.2              & 85.4  \\ 
\textbf{NCBID}        & 87.1               & 87.3                  & 88.3                   &  88.0              & 87.5              & 88.7 \\ 
\textbf{OntoNotes5*} & 84.9               & 85.2                  & 84.6                   &  84.8             & 85.4              & 84.8   \\ 
\textbf{WNUT2017}    & 49.5               & 53.3                  & 54.8                   &  52.1              & 53.5              & 55.0   \\ 
\textbf{ACE05-RE}    & 66.1               & 67.1                  & 68.0                   &   64.3             & 67.3              & 70.1   \\ 
\textbf{ACE05-EAE}    & 69.7               & 71.5                  & 72.7                   &   68.4             & 71.7              & 73.0   \\ 
\textbf{RAMS}        & 49.7               & 50.0                  & 49.9                   &   49.1           & 50.2              & 49.3    \\ 
\textbf{ACE05-EE}   & 70.1               & 70.3                  & 70.8                   &    68.5            & 70.5              & 71.8  \\ \hline
\textbf{Avg.Score}        & 76.0         & 76.8                 & \textbf{77.1}           &    75.7            & 77.0              & \textbf{77.6}              \\ \bottomrule[1.5pt]
\end{tabular}
\caption{We conducted extensive experiments on Qwen2.5-7B and QwenCoder2.5-7B to verify the generalization ability of MPL. We use Python in the Single PL setting in this experiment.}
\label{app:table.qwen}
\end{table*}

\begin{table*}[t]
\centering
\setlength{\tabcolsep}{1.5mm}
\begin{tabular}{c|ccc|cc|cc}
\toprule[1.5pt]
\textbf{}            & \textbf{Python}    & \textbf{C++}          & \textbf{Java}        & Python$_{3\times}$        & MPL$_{smapled}$  & \textbf{MPL} & w/o. Virtual Running  \\ \midrule[1.5pt]
\textbf{ACE05-NER}   & 89.68               & 90.04                  & 90.21                   &   90.98           & 90.33              & 91.40               & 90.49                                   \\ 
\textbf{BC5CDR}      & 89.06               & 88.77                  & 89.64                   &   88.38            & 89.86              & 89.62               & 89.19                                   \\ 
\textbf{CoNLL03}     & 93.13               & 92.73                  & 93.12                   &   92.80             & 93.46              & 93.54               & 93.28                                   \\ 
\textbf{DIANN}       & 82.22               & 82.94                  & 85.45                   &   83.81             & 85.52              & 85.39               & 88.34                                   \\ 
\textbf{NCBID}        & 87.61               & 86.82                  & 87.48                   &  87.74              & 88.49              & 88.06               & 87.54                                   \\ 
\textbf{OntoNotes5*} & 84.44               & 84.67                  & 84.63                   &  83.62             & 85.48              & 85.67                  & 85.43                                   \\ 
\textbf{WNUT2017}    & 53.59               & 53.18                  & 48.12                   &  52.75              & 54.58              & 52.58               & 54.87                                   \\ 
\textbf{ACE05-RE}    & 67.22               & 68.22                  & 69.47                   &   66.24             & 68.12              & 70.80               & 66.78                                   \\ 
\textbf{ACE05-EAE}    & 71.12               & 71.82                  & 70.83                   &   70.76             & 72.31              & 72.83               & 72.83                                  \\ 
\textbf{RAMS}        & 50.96               & 50.19                  & 50.05                   &   49.56           & 50.57              & 50.87                  & 49.97                                   \\ 
\textbf{ACE05-EE}   & 71.12               & 73.07                  & 71.84                   &    70.70            & 71.81              & 72.68               & 70.52                                   \\ \hline
\textbf{Avg.Score}        & 76.4      & 76.6   & 76.4                     &    76.1      & 77.1            & 77.6                  & 77.2                   \\ \bottomrule[1.5pt]
\end{tabular}
\caption{The detailed results of each model variant in the ablation study part, which is the supplementary of Section~\ref{sec:ablation}.}
\label{tab_app:ablation}
\end{table*}

